\documentclass[11pt,a4paper]{article}
\usepackage[hyperref]{acl2020}
\usepackage{times}
\usepackage{amsmath}
\usepackage{latexsym}
\usepackage{graphicx}

\usepackage{microtype}

\aclfinalcopy 

\newcommand{\short}{\textsc{MixingBoard}}

\newcommand{\reduceVerticalSpace}{
\setlength{\topsep}{0pt}
\setlength{\partopsep}{0pt}
\setlength{\parskip}{0pt}
\setlength{\parsep}{0pt}
\setlength{\itemsep}{0pt}
\setlength{\itemindent}{0pt}
\setlength{\listparindent}{0pt}
}

\title{\short:\\a Knowledgeable Stylized Integrated Text Generation Platform}

\author{Xiang Gao \quad\quad {Michel Galley} \quad\quad \textbf{Bill Dolan}\\
  Microsoft Research, Redmond, WA, USA \\
  {\small \tt \{xiag,mgalley,billdol\}@microsoft.com}
}

\date{}

\begin{document}
\maketitle
\begin{abstract}
We present \short, a platform for quickly building demos with a focus on knowledge grounded stylized text generation. We unify existing text generation algorithms in a shared codebase and further adapt earlier algorithms for constrained generation. To borrow advantages from different models, we implement strategies for cross-model integration, from the token probability level to the latent space level. An interface to external knowledge is provided via a module that retrieves on-the-fly relevant knowledge from passages on the web or any document collection. A user interface for local development, remote webpage access, and a RESTful API are provided to make it simple for users to build their own demos.
\footnote{Source code available at  \url{http://github.com/microsoft/MixingBoard}}. 
\end{abstract}


\section{Introduction}

\begin{figure}[t]
    \centering
    \includegraphics[width=0.47\textwidth]{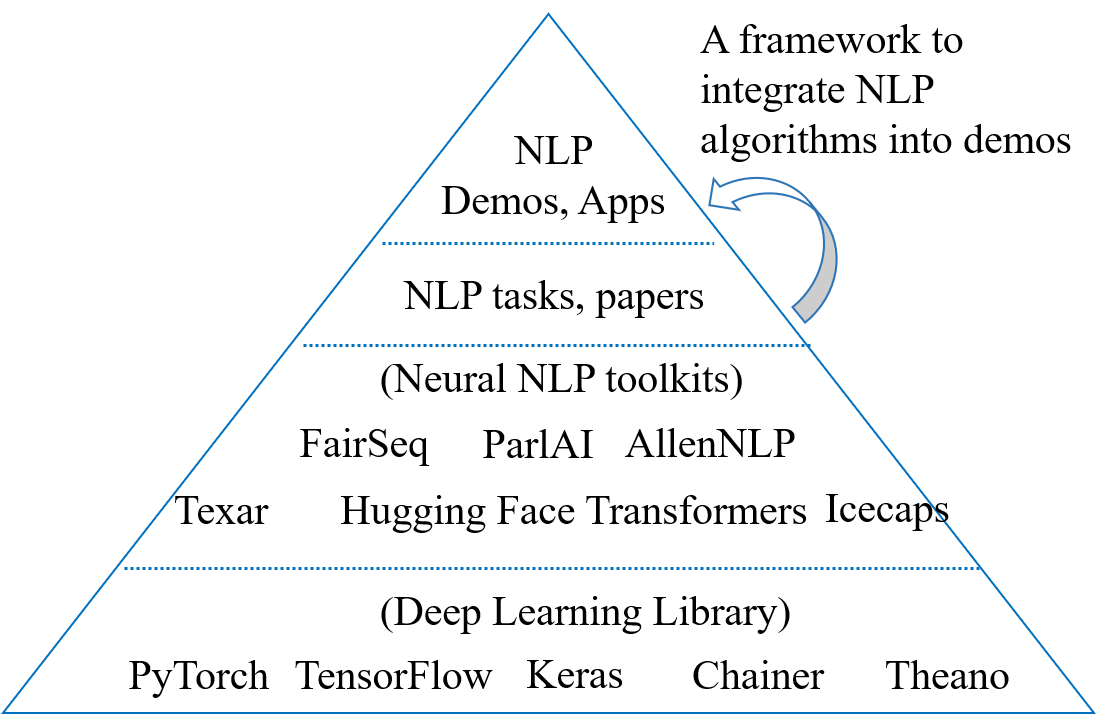}
    \caption{\short~is designed as a platform to organically and quickly integrate separate NLP algorithms into compelling demos}
    \label{fig:intro}
\end{figure}

Neural text generation algorithms have seen great improvements over the past several years \cite{radford2019language, gao2019neural}.
However each algorithm and neural model usually focuses on a specific task and may differ significantly from each other in terms of architecture, implementation, interface, and training domains. It is challenging to unify these algorithms theoretically, but a framework to organically integrate multiple algorithms and components can benefit the community in several ways, as it provides
(1) a shared codebase to reproduce and compare the state-of-the-art algorithms from different groups without time consuming trial and errors, (2) a platform to experiment the cross-model integration of these algorithms, and (3) a framework to build demo quickly upon these components. This framework can be built upon existing deep learning libraries \cite{paszke2019pytorch, tensorflow2015-whitepaper} and neural NLP toolkits \cite{pytorchtransformer, gardner2018allennlp, hu2018texar, ott2019fairseq, shiv2019microsoft, miller2017parlai}\footnote{Although multiple libraries and toolkits are mentioned in Fig.~\ref{fig:intro}, the current implementation is primarily based on PyTorch models}, as illustrated in Fig.~\ref{fig:intro}.
 
There are several challenges to do such integration. 
Firstly, engineering efforts are needed to unify the interface of different implementation.
Secondly, a top-level manager needs to be designed to utilize different models together. 
Finally, different models are trained using different data with different performance. Cross-model integration, instead of calling each isolated model individually, can potentially improve the overall performance.
In this work, we unified the models of different implementation in a single codebase, implemented demos as top-level managers to access different models, and provide strategies to allow more organic integration across the models, including token probability interpolation, cross-mode scoring, latent interpolation, and unified hypothesis ranking.

This work is also aimed to promote the development of grounded text generation. The existing works focusing on the knowledge grounded text generation \cite{prabhumoye2019towards, qin2019conversing, galley2019grounded, wu2020controllable} usually assume the knowledge passage is given. However in practice this is not true. We provide the component to retrieve knowledge passage on-the-fly from web or customized document, to allow engineers or researchers test existing or new generation models. Keyphrase constrained generation \cite{hokamp2017gbs} is another type of grounded generation, broadly speaking. Similarly the keyphrase needs to be provided to apply such constraints. We provide tools to extract constraints from knowledge passage or stylized corpus.

Finally, friendly user interface is a component usually lacking in the implementation of neural models but it is necessary for a demo-centric framework. We provide scripts to build local terminal demo, webpage demo, and RESTful API demo.

\section{Design}

\begin{figure*}[t]
    \centering
    \includegraphics[width=1\textwidth]{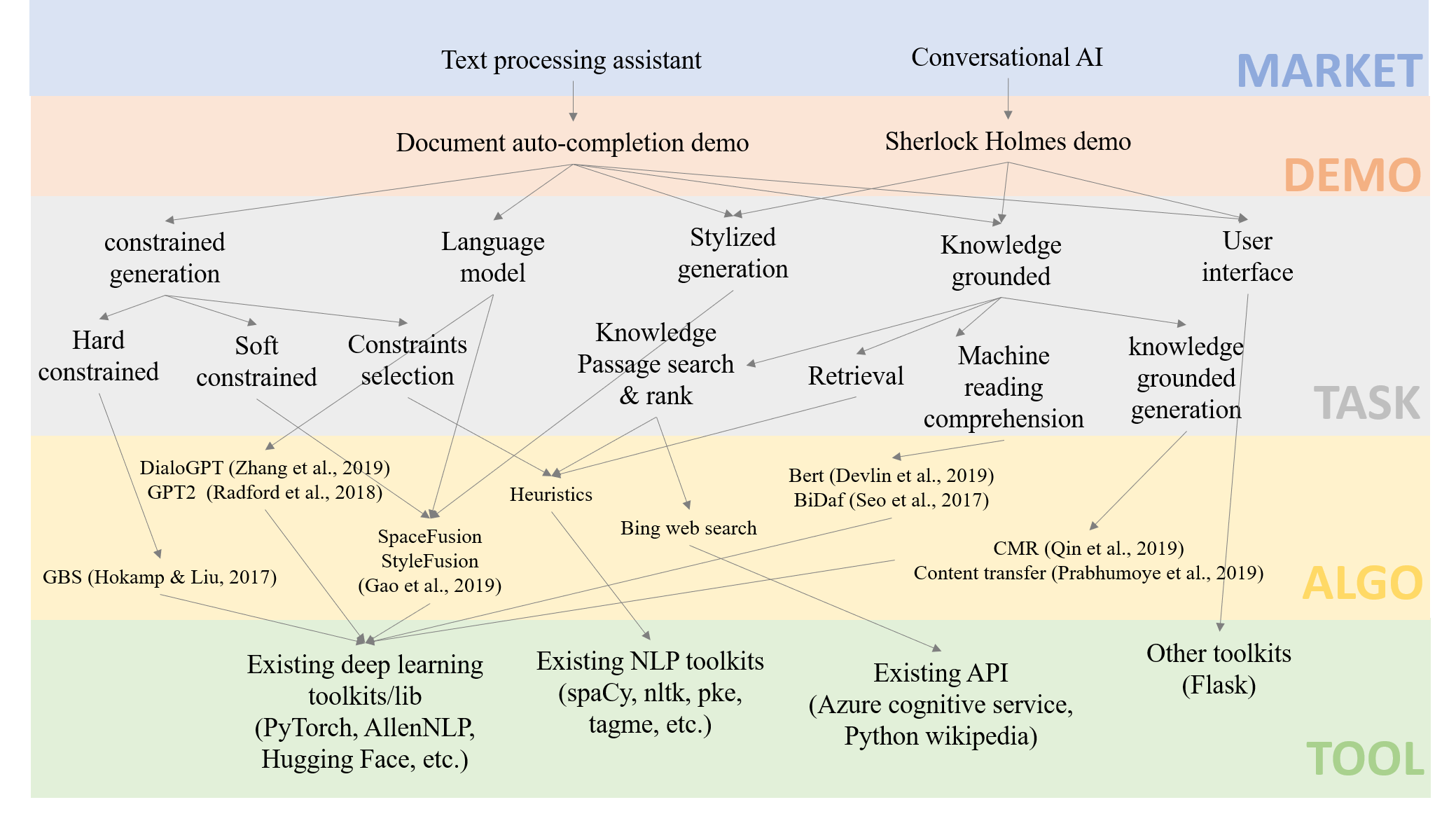}
    \caption{The architecture of \short, consisting of layers from basic tools, algorithms, tasks to integrated demos with market into consideration.}
    \label{fig:architecture}
\end{figure*}

Our goal is to build a framework that will allow users to quickly build text generation demos using existing modeling techniques.
This design allows the framework to be almost agnostic to the ongoing development of text generation techniques \cite{gao2019neural}. Instead, we focus on the organic integration of models and the interfaces for the final demo/app.

From a top-down view, our design are bounded to two markets: text processing assistant, and conversational AI, as illustrated in Fig.~\ref{fig:architecture}. Two demos are present as examples in these domains: document auto-completion and Sherlock Holmes. We further breakdown these demos into several tasks, designed to be shared across different demos. We also designed several strategies to integrate multiple models to generate text. These strategies allow each model to plug-in without heavy constraints on the architecture of the models, as detailed in Section~\ref{sec:integration}.

As the goal is not another deep learning NLP toolkit, we rely on existing ones \cite{pytorchtransformer, paszke2019pytorch, gardner2018allennlp} and online API services Bing Web Search provided in Azure Cognitive Service\footnote{\url{https://azure.microsoft.com/en-us/services/cognitive-services/}} and TagME.\footnote{\url{https://tagme.d4science.org/tagme/}} Similarly, most tasks are using existing algorithms: language modeling \cite{zhang2019dialogpt, radford2019language}, knowledge grounded generation \cite{qin2019conversing, prabhumoye2019towards} or span retrieval \cite{seo2016bidaf, devlin2018bert}, style transfer \cite{gao2019structuring, jointly2019gao} and constrained generation \cite{hokamp2017gbs}.  
\section{Modules}

\subsection{Knowledge passage retrieval}
We use the following free-text, unstructured text sources to retrieve relevant knowledge passage.

\begin{itemize}
\reduceVerticalSpace
   \item Search engine. Free-text form ``knowledge'' is retrieved from the following sources 1) text snippets from the (customized) webpage search; 2) text snippets (customized) news search; 3) user-provided documents. 
   \item Specialized websites. For certain preferred domains, e.g., wikipedia.org, we will further download the whole webpage (rather than just the text snippet returned from search engine) to obtain more text snippets.
   \item Users can also provide their customized knowledge base, like a README file, which can be updated on-the-fly, to allow the agent using latest knowledge.

\end{itemize}

User may select one or multiple sources listed above to obtain knowledge passage candidates. If the source does not provide a ranking of the snippets (e.g. paragraphs from a README file), then the text snippets are then ranked by relevancy to the query, measured by the keyphrases overlap between snippet. 

\subsection{Stylized synonym}

We provide a component to retrieve synonym of given target style for a query word. This component is useful for the style transfer module (Section~\ref{sec:stylized}) as well as the constrained generation module (Section~\ref{sec:constrained}).

The similarity based on word2vec, $\text{sim}_\text{word2vec}$, is defined as the cosine similarity between the vectors of two words. The similarity based on human-edited dictionary, $\text{sim}_\text{dict}$, is defined as 1 if the candidate word in the synonym list of the query word, otherwise 0. The final similarity between the two words is defined as the weighted average of these two similarities:
\begin{equation*}
\text{sim} = (1 - w_\text{dict}) \, \text{sim}_\text{word2vec} +  w_\text{dict} \, \text{sim}_\text{dict}
\end{equation*}
We only choose the candidate word with a similarity higher than certain threshold as the synonym of the query word. Then we calculate the style score of these synonym using a style classifier. We provided a logistic regression model taking 1gram multi-hot vector as features, trained on non-stylized corpus vs. stylized corpus.

\subsection{Latent interpolation}
\label{sec:interp}
For a given two latent vectors, $z_a$ and $z_b$, we expect the decoded results from the interpolated vector $z_i = u z_a + (1-u) z_b$ can retain the desired features from both $z_a$ and $z_b$. However this requires a interpolatable, smooth latent space. For this purpose, we apply the SpaceFusion \cite{jointly2019gao} and StyleFusion \cite{gao2019structuring, li2020optimus} to learn such latent space. The latent interpolation is then used to transfer style, apply soft constraints, and interpolating hypothesis obtained using different models.

\subsection{Stylized generation}
\label{sec:stylized}
\citet{gao2019structuring} proposed StyleFusion to generate stylized response for a given conversation context by structuring a shared latent space for non-stylized conversation data and stylized samples. We extend it to a style transfer method, i.e., modify the style of a input sentence while maintaining its content, via latent interpolation (see Section \ref{sec:interp}).

\begin{itemize}
\reduceVerticalSpace
   \item \textbf{Soft-edit} refers to a two-step algorithm, 1) edit the input sentence by replace each word by a synonym of the target style (e.g. ``cookie'' replaced by ``biscuit'' if the target style is British), if there exists any; 2) the edited sentence from step 1 may not be fluent, so we then apply latent interpolation between the input sentence and edited sentence to seek a sentence that is both stylized and fluent. 
   
   \item \textbf{Soft-retrieval} refers to a similar two-step algorithm, but step 1) is to retrieve a ``similar'' sentence from a stylized corpus, and then apply step 2) to do the interpolation. One example is given in Fig.~\ref{fig:web-sherlock}. The hypothesis ``he was once a schoolmaster in the north of england'' is retrieved given the DialoGPT hypothesis ``he's a professor at the university of london''.
\end{itemize}

\subsection{Conditioned text generation}

Generate a set of candidate responses conditioned on the conversation history, or a follow-up sentence conditioned on the existing text.

\begin{itemize}
\reduceVerticalSpace
   \item GPT-2 \cite{radford2019language} is a transformer \cite{vaswani2017attention} based text generation model.
   
   \item DialoGPT \cite{zhang2019dialogpt} is a large-scale pre-trained conversation model obtained by training GPT-2 \cite{radford2019language} on Reddit comments data.
   
   \item SpaceFusion \cite{jointly2019gao} is a regularized multi-task learning framework proposed to learn a smooth and interpolatable latent space.
\end{itemize}

\subsection{Knowledge grounded generation}
We consider the following methods to consume the retrieved knowledge passage and
relevant long-form text on the fly as a
source of external knowledge.

\begin{itemize}
\reduceVerticalSpace
   \item Machine reading comprehension. In the codebase, we fine-tuned BERT on SQuAD.
   
   \item Content transfer is a task proposed in \cite{prabhumoye2019towards} designed to, given a context, generate a sentence using knowledge from an external article. We implemented this algorithm in the codebase.
   
   \item Knowledge grounded response generation is a task firstly proposed in \cite{ghazvininejad2018knowledge} and later extended in Dialog System Technology Challenge 7 (DSTC7)\cite{galley2019grounded}. We implemented the CMR algorithm (Conversation with on-demand Machine Reading) proposed in \cite{qin2019conversing}.
\end{itemize}

\subsection{Constrained generation}
\label{sec:constrained}
Besides the grounded generation, it is also useful to apply constraints at the decoding stage that encourage the generated hypotheses contain the desired phrases. 
We provide the following two ways to obtain constraints.

\begin{itemize}
\reduceVerticalSpace
   \item Key phrases extracted from the Knowledge passage. We use the PKE package \cite{boudin2016pke} to identify the keywords.
   \item In some cases, users may want to use a stylized version of the topic phrases or phrase of a desired style as the constraints. We use the stylized synonym algorithm as introduced in Section~\ref{sec:stylized} to provide such stylized constraints.
\end{itemize}

With the constraints obtained above, we provide the following two ways to apply such constraints during decoding.

\begin{figure}[t]
    \centering
    \includegraphics[width=0.5\textwidth]{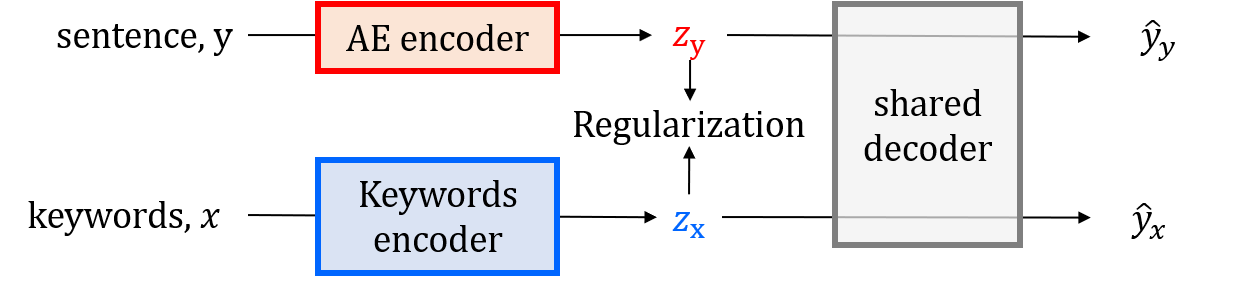}
    \caption{A soft keywords constrained generation model based on SpaceFusion \cite{jointly2019gao}.}
    \label{fig:keywords-fusion}
\end{figure}

\begin{itemize}
\reduceVerticalSpace
   \item \textbf{Hard constraint} is applied via Grid Beam Search (GBS) \cite{hokamp2017gbs}, which is a lexically constrained decoding algorithm that can be applied to almost any models at the decoding stage and generate hypotheses that contain desired paraphrases (i.e. the constraints). We implemented GBS to provide a hard constrained decoding.
   \item \textbf{Soft constraint} refers the case that generation is likely, but not always, to satisfy constraints (e.g. include given keywords in the hypothesis). We provide an adapted version of SpaceFusion \cite{jointly2019gao} for this purpose. \citet{jointly2019gao} proposed to align the latent space of a Sequence-to-Sequence (S2S) model and that of an Autoencoder (AE) model to improve dialogue generation performance. Inspired by this work, we proposed to replace the S2S model by a keywords-to-sequence model, which takes multi-hot input of the keywords $x$ identified from sentence $y$, as illustrated in Fig.~\ref{fig:keywords-fusion}. During training, we simply choose the top-k rare words (rareness measured by inverse document frequency) as the keywords, and k is randomly choose from a Uniform distribution $k \sim U(1,K)$. 
\end{itemize}

\section{Cross-model integration}
\label{sec:integration}

\begin{figure*}[t]
    \centering
    \includegraphics[width=1\textwidth]{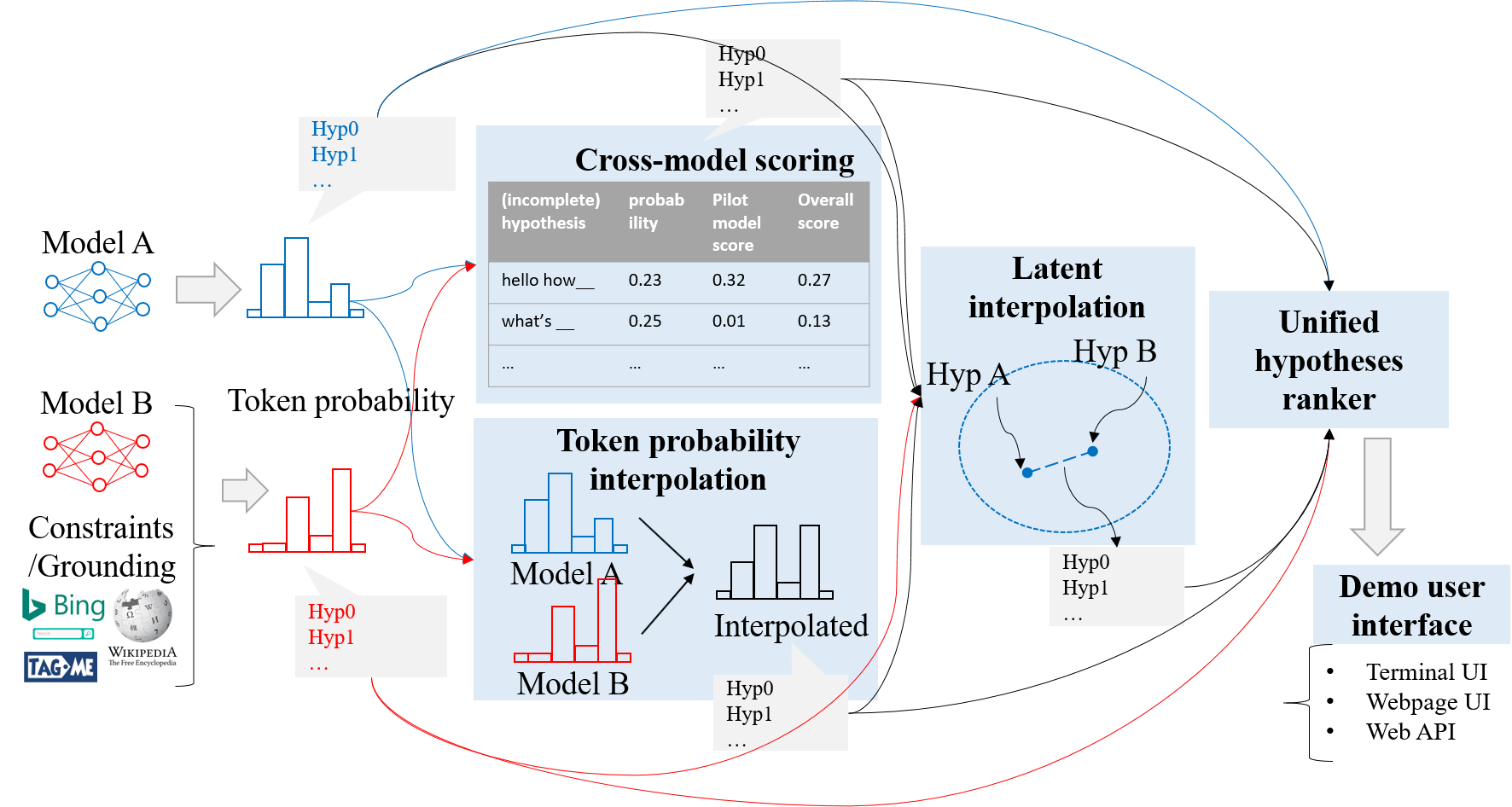}
    \caption{An example flow chart showing the integration of two models at different stages (blue boxes).}
    \label{fig:integration}
\end{figure*}

Multiple models may be called for the same query and returns different responses. We propose the following ways to organically integrate multiple models, as illustrated in Fig.~\ref{fig:integration}. User can apply these strategies with customized models.

\begin{itemize}
\reduceVerticalSpace
    \item \textbf{Token probability interpolation} refers prediction of the next token using a (weighted) average of the token probability distributions from two or more models given the same time step and given the same context and incomplete hypothesis. Previously, it has been proposed to bridge a conversation model and stylized language model \cite{niu2018polite}. This technique does not require the models to share the latent space but the vocabulary should be shared across different models.
    
    \item \textbf{Latent interpolation} refers the technique introduced in Section~\ref{sec:interp}. It provides a way to interpolate texts in the latent space. Unlike the token-level strategy introduced above, this technique focuses on the latent level and ingests information from the whole sentence. However if the two candidates are too dissimilar, the interpolation may result in undesired outputs. The soft constraint algorithm introduced in Section~\ref{sec:constrained} is one option to apply such interpolation.
    
    \item \textbf{Cross model pruning} refers pruning the hypothesis candidates (can be incomplete hypothesis, e.g. during beam search) not just based on the joint token probability, but also the evaluated probability from a secondary model. 
    This strategy does not require a shared vocabulary or a shared latent space. Interpolating two models trained on dissimilar domains may be risky but the cross model pruning strategy is safer as the secondary model is only used roughly as a discriminator rather than a generator.
    
    \item \textbf{Unified hypothesis ranking} is the final step which sum up the hypotheses generated from each single model and these from the integration of multiple models using the above strategies. We consider the following criteria for the hypothesis ranking: 1) likelihood, measured by the conditional token probability given the context; 2) informativeness, measured by average inverse document frequency (IDF) of the tokens in the hypothesis; 3) repetition penalty, measured by the ratio of the number of unique ngrams and the number of total ngrams. and 4) style intensity, measured by a style classifiers, if style is considered.
    
\end{itemize}

\section{Demos}
\label{sec:demos}

\subsection{Virtual Sherlock Holmes}

This demo is a step towards a virtual version of Sherlock Holmes, able to chat in Sherlock Holmes style, with Sherlock Holmes background knowledge in mind. As an extended version of the one introduced by \citet{gao2019structuring}, the current demo is grounded on knowledge and coupled with more advanced language modeling \cite{zhang2019dialogpt}.
It is designed to integrate the following components: open-domain conversation, stylized response generation, knowledge-grounded conversation, and question answering. Specifically, for a given query, the following steps are executed:
\begin{itemize}
\reduceVerticalSpace
  \setlength\itemsep{0em}
   \item Call DialoGPT \cite{zhang2019dialogpt} and StyleFusion \cite{gao2019structuring} to get a set of hypotheses.
   
   \item Call the knowledge passage selection module to get a set of candidate passages. Then feed these passages to the span selection algorithm (Bert-based MRC \cite{devlin2018bert}) and CMR \cite{qin2019conversing} to get a set of knowledge grounded response. 
   
   \item Optionally, use the cross-model integration strategies, such as interpolating the token probability of DialoGPT and CMR.
   
   \item Based on TF-IDF similarity, best answer is retrieved from a user provided corpus of question-answer pairs. If the similarity is lower than certain threshold, the retrieved result will not be returned.
   
   \item Apply the style transfer module to obtain stylized version of the the hypotheses obtained from steps above.
   
   \item feed all hypotheses to the unified ranker and return the top ones.
   
\end{itemize}

\subsection{Document auto-completion assistant}

This demo is designed as a writing assistant, which provides suggestion of the next sentence given the context. The assistant is expected to be knowledgeable (able to retrieve relevant knowledge passage from web or a given unstructured text source) and stylized (if a given target style is specified). For a given query, the following steps are executed:
\begin{itemize}
\reduceVerticalSpace
   \item Call language model GPT2 \cite{radford2019language} to get a set of hypotheses
   
   \item Call the knowledge passage selection module to get a set of candidate passages. Then feed these passages to content transfer algorithm \cite{prabhumoye2019towards} to get a set of knowledge grounded response. 
   
   \item Optionally, use the cross-model integration strategies, such as latent interpolation to interpolate hypotheses from above models. 
   
   \item Apply the style transfer module to obtain stylized version of the the hypotheses obtained from steps above.
   
   \item Feed all hypotheses to the unified ranker and return the top ones.
   
\end{itemize}

\section{User interface}

We provided the following three ways to access the demos introduced above for local developer, remote human user, and interface for other programs.

\begin{figure}[t]
    \centering
    \includegraphics[width=0.48\textwidth]{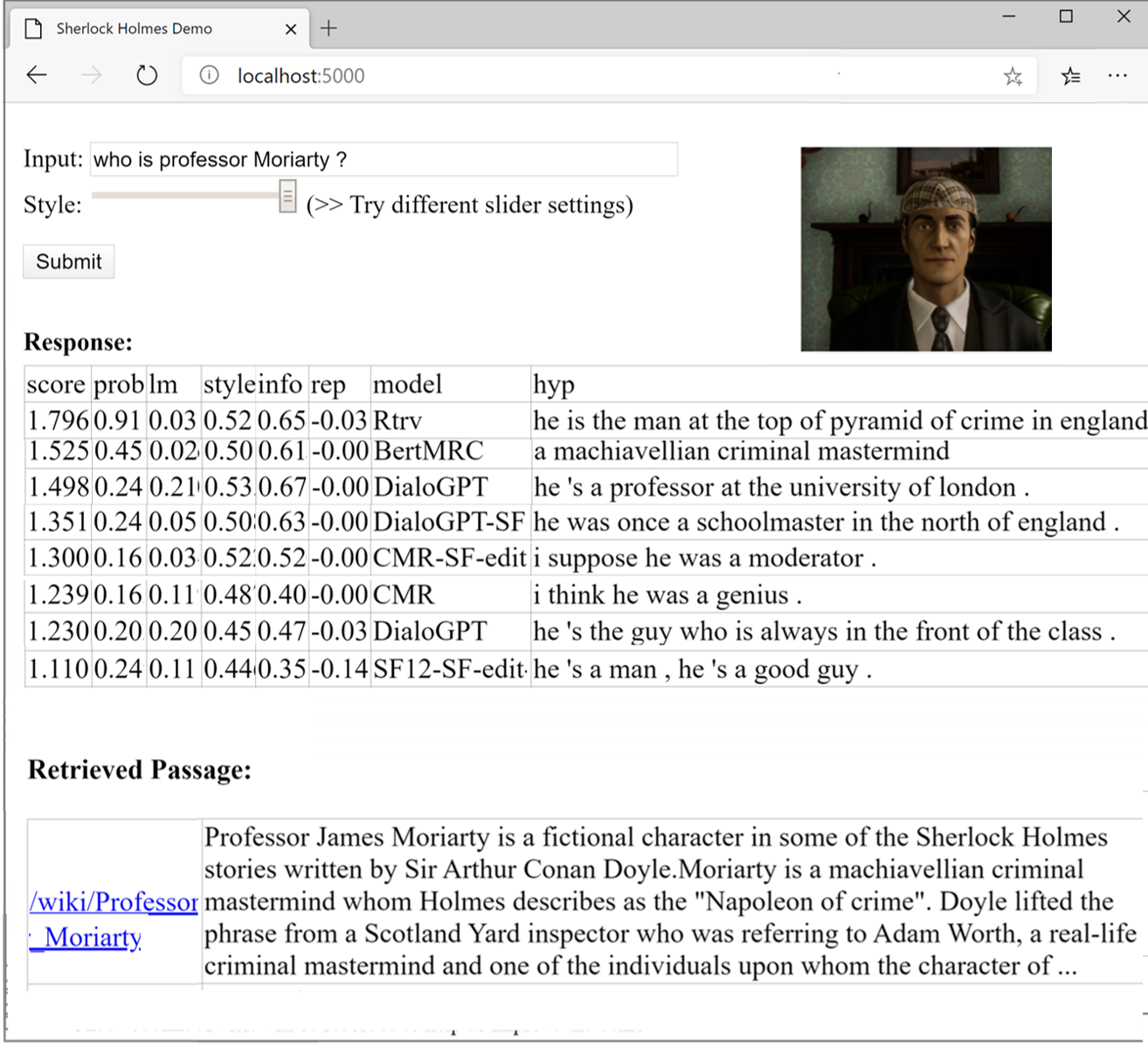}
    \caption{Sherlock Holmes webpage demo with wikipedia knowledge example. 
    }
    \label{fig:web-sherlock}
\end{figure}

\begin{figure}[t]
    \centering
    \includegraphics[width=0.48\textwidth]{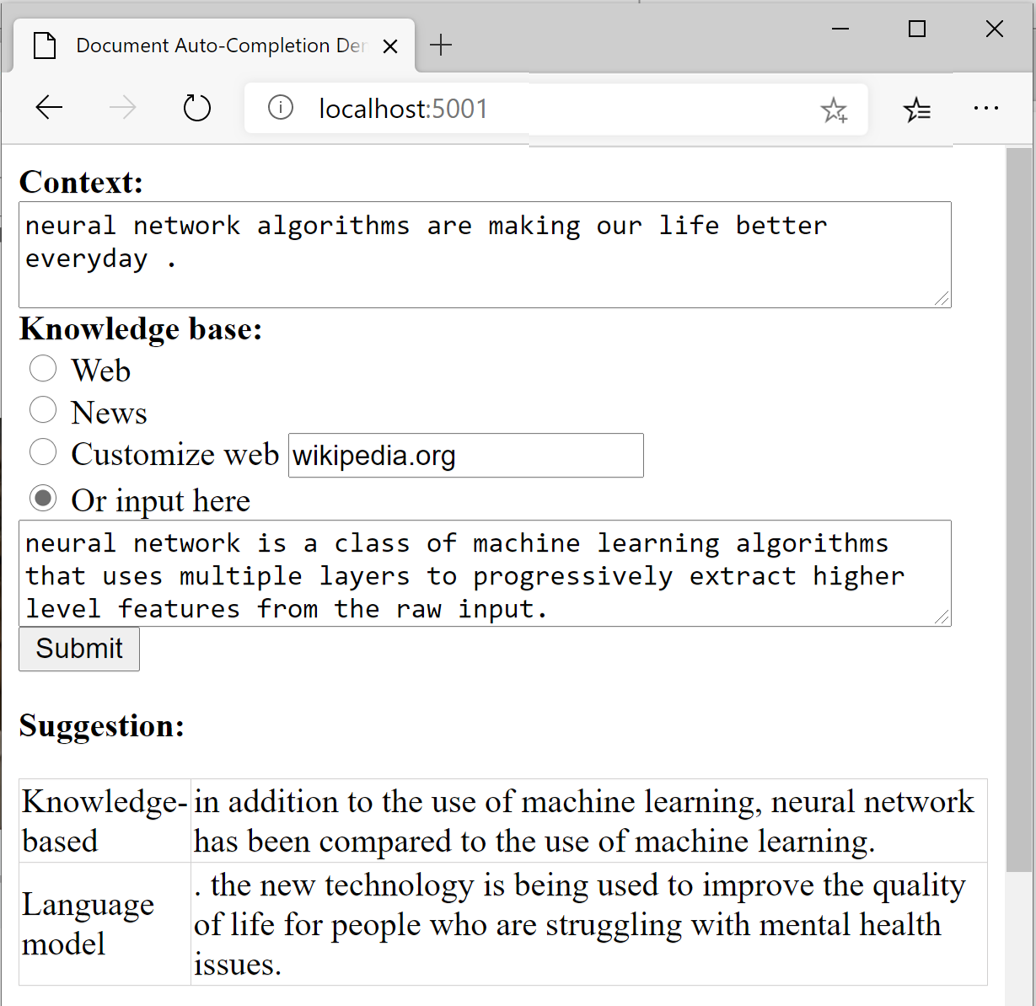}
    \caption{Document Auto-completion webpage demo with user input knowledge passage.}
    \label{fig:web-dac}
\end{figure}

\begin{itemize}
  \setlength\itemsep{0em}
   \item \textbf{Command line interface} is provided for local interaction. This is designed for developer to test the codebase.
   
   \item \textbf{Webpage interface} is implemented using the Flask toolkit.\footnote{\url{https://flask.palletsprojects.com/en/1.1.x/}} A graphic interface is provided with html webpage for remote access for human user. As illustrated in Fig.~\ref{fig:web-sherlock}, the Sherlock Holmes webpage consists of a input panel where the user can provide context and control style, a hypothesis list which specify the model and scores of the ranked hypotheses, and a knowledge passage list showing the retrieved knowledge passages. Another example is given in Fig.~\ref{fig:web-dac} for document auto-completion demo, where multiple options of knowledge passage is given.
   
   \item \textbf{RESTful API} is implemented using the Flask-RESTful toolkit.\footnote{\url{https://flask-restful.readthedocs.io/en/latest/}} JSON object will be returned for remote request. This interface is designed to allow remote access for other programs. One example is to host this RESTful API on a dedicated GPU machine, so the webpage interface can be hosted on another less powerful machine to send request through RESTful API. 
   
\end{itemize}
\section{Conclusion}

\short~is a new open-source platform to organically integrate multiple state-of-the-art NLP algorithms to build demo quickly with user friendly interface. 
We unified these NLP algorithms in a single codebase, implemented demos as top-level managers to access different models, and provide strategies to allow more organic integration across the models.
We provide the component to retrieve knowledge passage on-the-fly from web or customized document for grounded text generation.
For future work, we plan to keep adding the state-of-the-art algorithms, reduce latency and fine-tune the implemented models on larger and/or more comprehensive corpus to improve performance.

\bibliography{ref}
\bibliographystyle{acl_natbib}

\end{document}